\documentclass{bmvc2k}

\title{Humans need not \textit{label} more humans:\\ Occlusion Copy \& Paste for Occluded Human Instance Segmentation}

\addauthor{Evan Ling}{evan.ling@hmgics.com}{1}
\addauthor{Dezhao Huang}{dezhao.huang@hmgics.com}{1}
\addauthor{Minhoe Hur}{minhoe.hur@hmgics.com}{1}

\addinstitution{
 AIR Center\\
 Hyundai Motor Group Innovation Center in Singapore\\
 Singapore
}

\runninghead{Ling, Huang, Hur}{Occlusion Copy \& Paste}

\def\eg{\emph{e.g}\bmvaOneDot}

\def\ie{\emph{i.e}\bmvaOneDot}

\begin{document}

\maketitle
\begin{abstract}

Modern object detection and instance segmentation networks stumble when picking out humans in crowded or highly occluded scenes. Yet, these are often scenarios where we require our detectors to work well. Many works have approached this problem with model-centric improvements. While they have been shown to work to some extent, these supervised methods still need sufficient relevant examples (\ie occluded humans) during training for the improvements to be maximised. In our work, we propose a simple yet effective data-centric approach, Occlusion Copy \& Paste, to introduce occluded examples to models during training --- we tailor the general copy \& paste augmentation approach to tackle the difficult problem of same-class occlusion. It improves instance segmentation performance on occluded scenarios for ``free'' just by leveraging on existing large-scale datasets, without additional data or manual labelling needed. In a principled study, we show whether various proposed add-ons to the copy \& paste augmentation indeed contribute to better performance. Our Occlusion Copy \& Paste augmentation is easily interoperable with any models: by simply applying it to a recent generic instance segmentation model without explicit model architectural design to tackle occlusion, we achieve state-of-the-art instance segmentation performance on the very challenging \emph{OCHuman} dataset. \emph{Source code is available at \url{https://github.com/levan92/occlusion-copy-paste}.}

\end{abstract}

\section{Introduction}
\label{sec:intro}

\begin{figure}
\centering
\begin{tabular}{cc}
\includegraphics[width=0.47\textwidth]{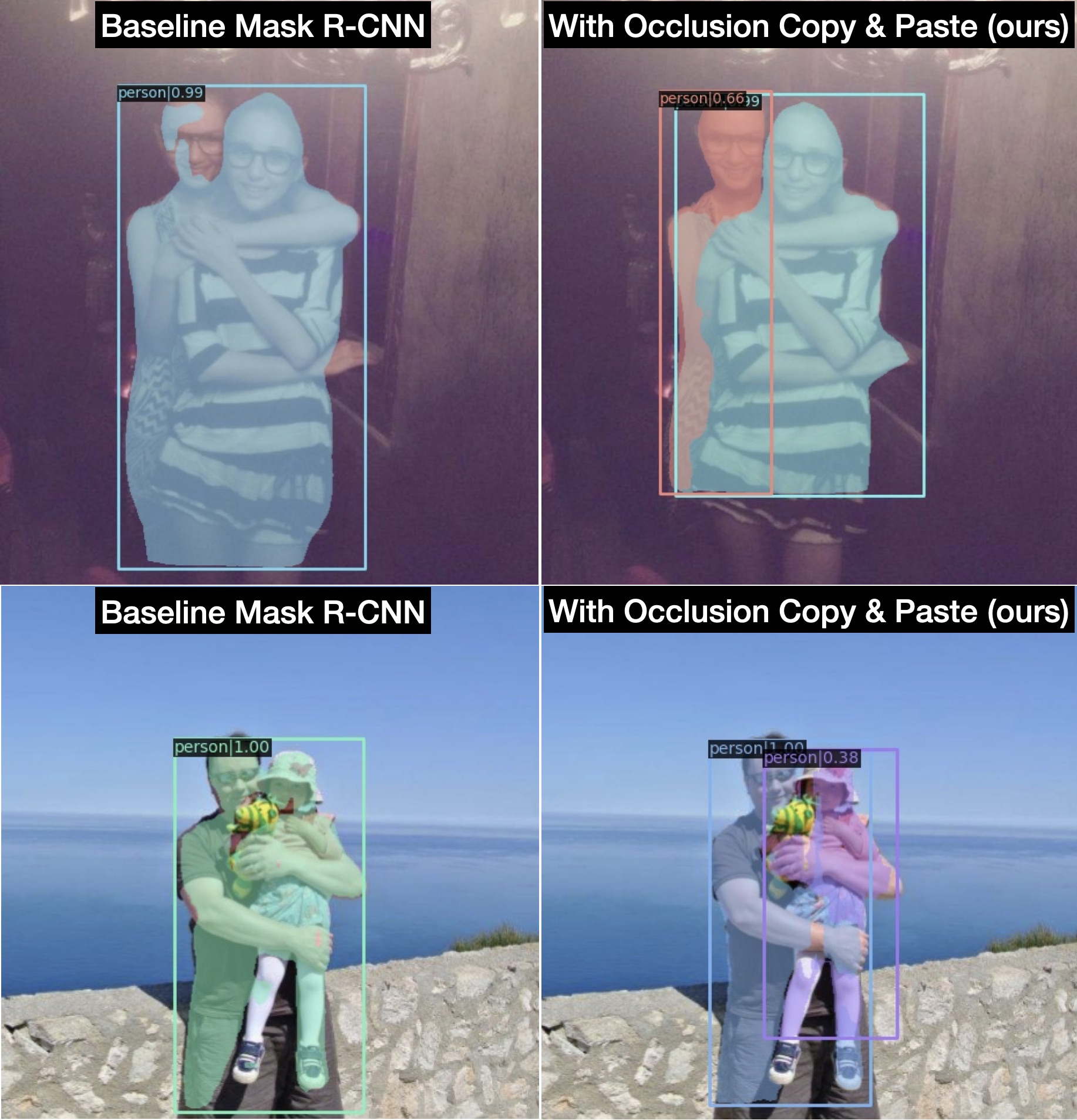} & 
\includegraphics[width=0.43\textwidth]{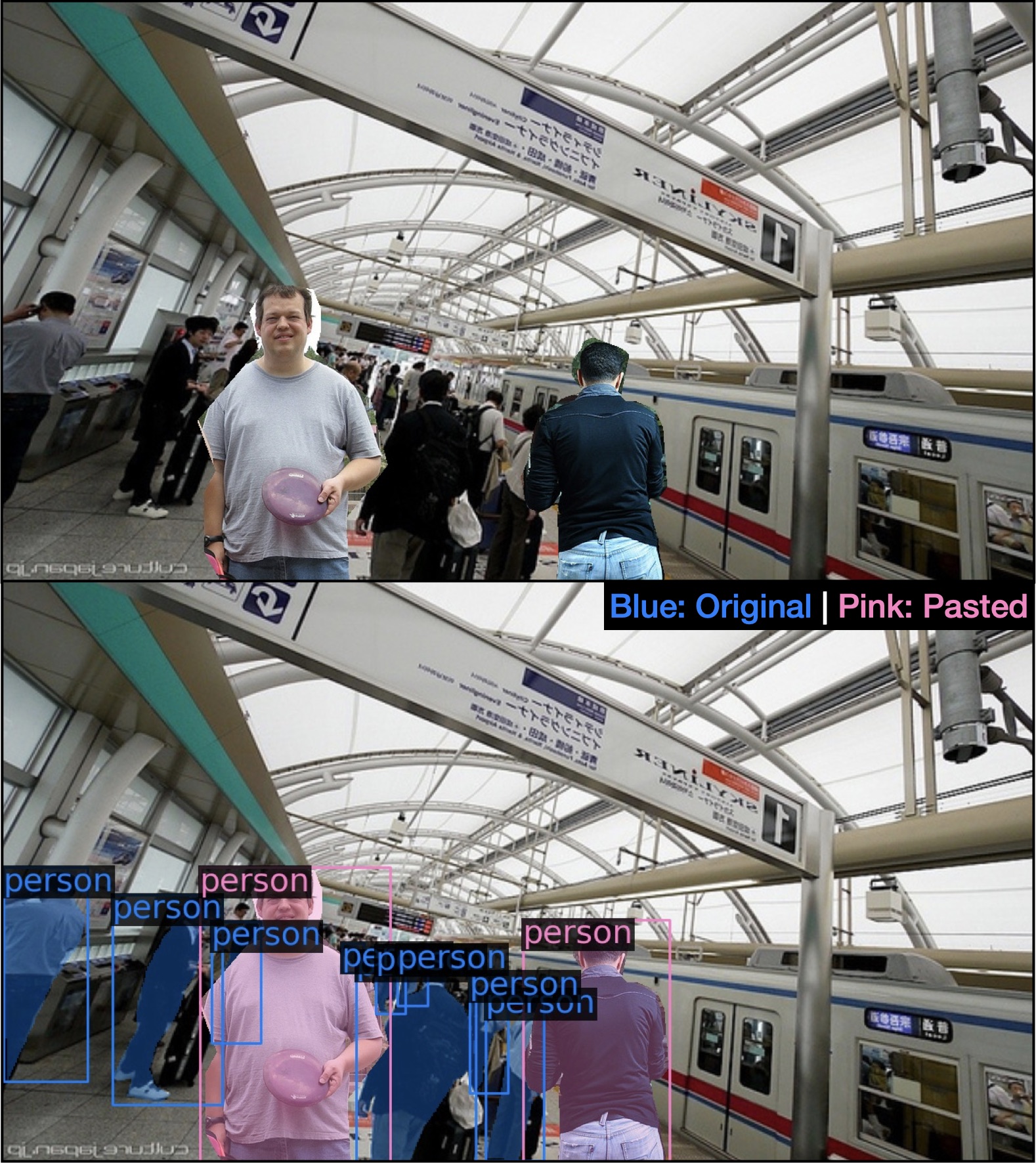} \\
(a) & (b)
\end{tabular}
\caption{(a) Instance segmentation predictions of Mask R-CNN on occluded human cases versus one trained with our Occlusion Copy \& Paste (base images from \emph{OCHuman} \cite{zhang2019pose2seg}); \\
(b) Example of image augmented with our Occlusion Copy \& Paste approach (base image from MS COCO dataset \cite{lin2014microsoft})}\label{fig:compare_example}
\vspace{-5mm}
\end{figure}

There is nothing more interesting to humans, than humans. Object-level visual recognition tasks like object detection and segmentation of the \emph{person} class have overarching applications from pedestrian detection for autonomous driving, to worker safety monitoring in factories and construction sites, to retail analytics in the commercial space. The fact that humans are everywhere necessitates that we need to get very good at picking out humans. However, current state-of-the-art (SOTA) object detection and instance segmentation models often fail when people are occluded and in crowded scenes \cite{wang2018repulsion,zhang2019pose2seg, ke2021deep}. The issue is exacerbated in instance segmentation when we need to delineate multiple neighboring instances of the same class occluding each other. The current model performance on such cases is illustrated on the left of \Cref{fig:compare_example}(a). Our work looks at this harder problem of same-class occlusion in instance segmentation, and more specifically, of the \emph{person} class, the most common class of instances in the MS COCO dataset \cite{lin2014microsoft}. Moreover, as discussed above, there are ubiquitous applications for recognizing humans. 

Methods to tackle most problems in deep learning can be categorized into model-centric or data-centric approaches --- most of the time, people found most success using a combination of both. In the space of occluded instance recognition, good model-centric approaches have been proposed to tackle occlusions \cite{wang2018repulsion, zhang2019pose2seg, ke2021deep, lazarow2020learning}. However, most of these are supervised, inherently assuming availability of relevant labelled data to learn from, \ie, labelled images with instance occlusions. While it has been very promising in the self-/semi-supervised space \cite{he2020momentum, chen2020simple, tarvainen2017mean}, the current state of deep learning is still very much label-hungry. Nothing improves a model more than simply having many relevant labelled data.

It is commonplace for detection \& segmentation models to be trained (or pre-trained) on open-source large-scale datasets\cite{everingham2010pascal, lin2014microsoft}. However, such datasets comprise of common scenes from daily life and difficult scenes with occlusions are few and far between. This possibly contributes to the poor performance of modern models on occluded scenes. A naïve way to go about this is to collect and label a sufficiently large dataset of occluded people. Unfortunately, segmentation tasks are notoriously hard to label for. It took workers an average of 79.2 seconds per instance to annotate the segmentation masks in the MS COCO dataset \cite{lin2014microsoft} ---  a total of 2,500,000 instances took 55,000 man-hours. The effort will be prohibitively large to collect and annotate a large-scale dataset of occluded humans, given that the labelling effort would be even harder due to occluded \& intertwined instances.

In light of these limitations, we propose the novel use of instance-level copy \& paste as a form of augmentation to directly induce occlusion during training. \Cref{fig:compare_example}(b) shows an example training image after applying our augmentation. With our proposed Occlusion Copy \& Paste, this straightforward approach significantly improves instance segmentation performance on highly occluded human scenarios (example results on right of \Cref{fig:compare_example}(a)), all whilst training on existing datasets without additional data collection or labelling. For the first time, to the best of our knowledge, copy \& paste is used as an augmentation to directly tackle the issue of occluded instance segmentation. We studied the necessity of realism in augmented images and found that most of the time, it is actually counter-productive. We also experimented and explored the space of using copy \& paste to refine its effectiveness as an augmentation technique to tackle occluded instance segmentation. Our approach is complementary to, and should be coupled with model-centric approaches in tackling occlusions to achieve the best performance.  

To summarise, our key contributions are as follows: \textbf{(1)} We propose the novel use of instance-level copy \& paste augmentation to tackle the problem of occluded person instance segmentation; \textbf{(2)} We conducted a principled study on the efficacy of our various add-ons that tailors our copy \& paste augmentation for occluded instance segmentation and importantly show that sometimes, variety is favoured over realism; \textbf{(3)} By simply applying our Occlusion Copy \& Paste with a recent instance segmentation model \cite{cheng2022masked}, without any explicit model architectural designs to tackle occlusions, we achieve SOTA instance segmentation performance on the very challenging \emph{OCHuman} benchmark \cite{zhang2019pose2seg}; \textbf{(4)} We also clarify various labelling and evaluation fairness issues around \emph{OCHuman} and propose a fully-labelled subset for future works to benchmark upon: \emph{OCHuman\textsuperscript{FL}}. 

\section{Related Work}
\label{sec:related}

\paragraph{Object Detection \& Instance Segmentation.}\label{par:related_od}Object detection and instance segmentation are closely linked in concept and research works. Modern object detection is anchored upon the seminal R-CNN work \cite{girshick2014rich} in 2014 and its succeeding Fast \& Faster R-CNN \cite{girshick2015fast, ren2015faster} which formed the core framework for two-stage object detection for nearly the decade to come. Two-stage models first propose regions of interest, then refine bounding boxes and classify in the second stage. For instance segmentation, Mask R-CNN \cite{he2017mask} was a natural extension of Faster R-CNN \cite{ren2015faster}, extending a mask head parallel to the existing box and classification heads. This completely shifted how instance segmentation is approached, bringing forth an “instance-first” strategy, which proved to be highly effective despite being conceptually simple. Most recently, Mask2Former \cite{cheng2022masked} was introduced as a universal model architecture for segmentation tasks, which broke the SOTA in not just instance but across semantic and panoptic segmentation as well. Using a DETR \cite{carion2020end} style transformer-based decoder with masked attention and learnable object queries allowed for efficient training while exploiting the pair-wise self-attention interaction in the transformer layers. The goal of our work is to prove the efficacy of copy \& paste augmentation in tackling the issue of occlusions in instance segmentation. Therefore, we chose Mask R-CNN \cite{he2017mask} as our baseline model during experimentation due to its simplicity and efficiency. We eventually extend our approach to Mask2Former \cite{cheng2022masked}, demonstrating SOTA results on occluded human benchmarks. \\
\vspace{-6mm}

\paragraph{Tackling Occlusions.} \label{par:related_occ} Occlusion is not a new problem --- many works over the years have attempted to tackle it from a model-centric perspective, from occlusion in object detection \cite{wang2018repulsion, zhang2018occlusion} to segmentation like OCFusion \cite{lazarow2020learning} and BCNet \cite{ke2021deep}. More relevant to our work, \cite{zhang2019pose2seg} introduced a new benchmark for occluded person instance segmentation, the Occluded Human (\emph{OCHuman}) dataset. This challenging benchmark is especially useful, plugging the gap on instance segmentation datasets containing heavily occluded humans. In the same work, they propose to tackle occlusion via Pose2Seg, where instead of bounding boxes, human poses are used as proposals to instance segmentation and used for feature alignment. Following, PoSeg \cite{zhou2020poseg} extends it by integrating top-down and bottom-up cues, predicting human pose and instance segmentation mask in a multi-task manner. Similarly, they benchmark their approach on the \emph{OCHuman} dataset. Most works have tackled occlusion with supervised model-centric approaches. Complementing that, our work approaches the occlusion issue from a data-centric direction by providing relevant labelled data. \\
\vspace{-7mm}

\paragraph{Copy \& Paste as Augmentation.} \label{par:related_cp} Image augmentation regularizes training by introducing greater variety of training inputs. More complex forms of augmentation bring in multiple images, like mixup \cite{zhang2017mixup}, CutMix \cite{yun2019cutmix} or mosaic augmentation of Yolov4 \cite{bochkovskiy2020yolov4}. These ideas run along the same vein as the concept of copy \& pasting instances onto another image. This copy \& paste concept is simple and over the years, several works have tried to formalize the idea, proposing variations for different uses. One of the earlier works \cite{dwibedi2017cut} used it for instance detection: from a separate bank of segmented instances, they pasted instances randomly onto backgrounds, framed as a synthetic data approach instead of as an augmentation technique. They particularly found that their model was over-fitting to pasting artefacts at the edges and implements random blending for the model to look beyond these artefacts and blending styles. A subsequent work by \cite{dvornik2018modeling} argues that random pasting does not extend to object detection, proposing an additional contextual model to determine if an instance is of suitable class for pasting into an image. Then, InstaBoost \cite{fang2019instaboost} eliminates the extra contextual model, making use of original instances within an image and jitter them around their locality instead.

Most recently, \cite{ghiasi2021simple} primes Simple Copy-Paste (SCP) for improving instance segmentation performance. This work surprisingly rebuts the various claims from the previous works \cite{dwibedi2017cut, dvornik2018modeling}---they claim that blending is not beneficial and that pasting with context is not needed, random pasting works just as well. Similarly in our experiments, we show that such realism enhancers are actually counter-productive most of the time. SCP views copy \& paste as a way to generally increase diversity of data points during training, simply taking a pair of images and directly overlaying instances from one image over the other without modifications. In contrast, we use copy \& paste as a means to directly induce occlusions and at the same time retain “free” ground-truth labels via Occlusion Copy \& Paste. We introduce more direction and stochasticity in pasting---copy \& paste is used as image augmentation to directly tackle the issue of occluded persons instance segmentation.

\section{Approach}
\label{sec:approach}

The concept behind copy \& paste is straightforward: we start from our Basic Copy \& Paste pipeline, then introduce some add-ons and designs that improves efficiency, stochasticity and enhance realism. We show in experiments later that realism enhancements are generally counter-productive. The eventual Occlusion Copy \& Paste does not contain most of the realism enhancements, but do contain the add-ons that improve efficiency and stochasticity.

\begin{figure}
\centering
\includegraphics[width=0.7\textwidth]{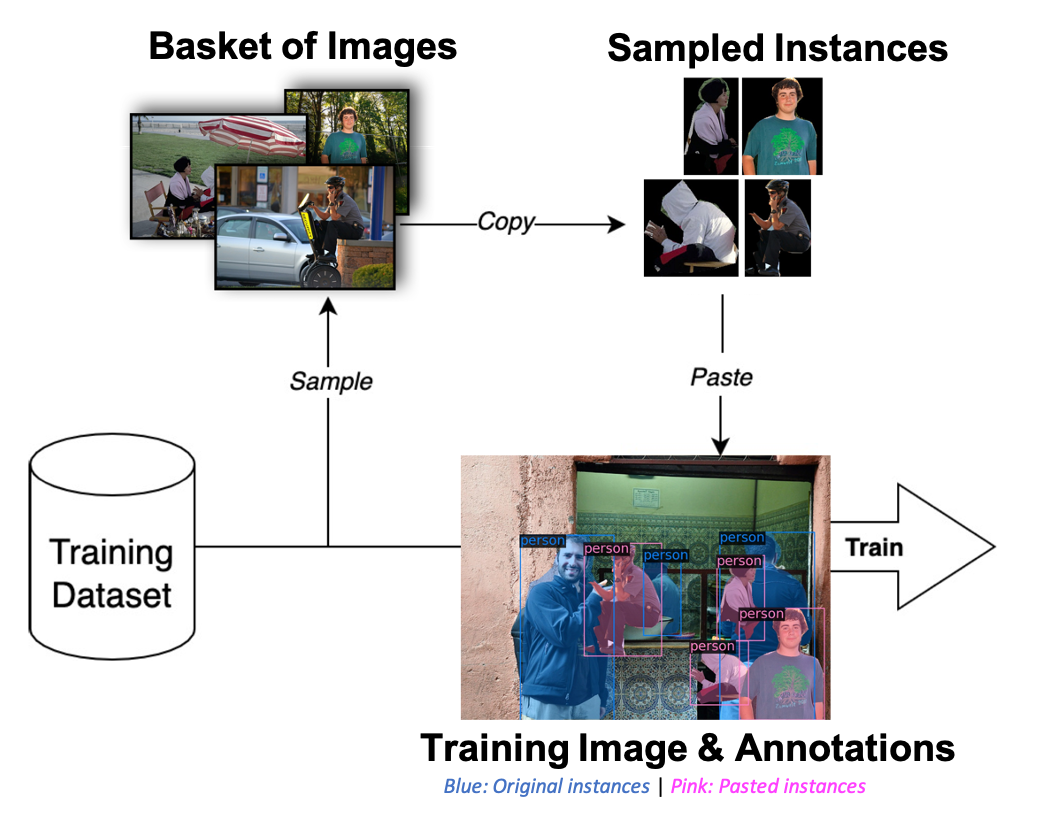}
\vspace*{-5mm}
\caption{Overview of our Basic Copy \& Paste augmentation pipeline.}\label{fig:approach}
\vspace{-3mm}
\end{figure}

\subsection{Basic Copy \& Paste}

Figure \ref{fig:approach} illustrates our Basic Copy \& Paste pipeline for each data iteration during training: For every training image, a dynamic basket of images is sampled from the dataset. Then, a random subset of instances from this basket is pasted onto the current image at random locations. Sub-regions of existing instances that overlapped with pasted instances may no longer be visible, so the associated ground-truth mask regions are removed. Some instance may be too occluded to meaningfully enforce predictions: if remaining visible regions is too small, we will remove that instance entirely. Instances are pasted in sequentially, which facilitates the possibility of occlusions amongst pasted instances as well. This runs in real-time in each training iteration, giving rise to novel views \& occlusion scenarios in every step, the combinatorial sum of all instances and images provides relevant yet huge variety of training examples to the task of occluded instance segmentation. 

The augmentation pipeline has the following hyper-parameters introduced:
\begin{itemize}[noitemsep,topsep=0pt]
  \item $\mathbf{P}_{CP}$: Overall probability for Copy \& Paste augmentation, sampled every iteration, ensures it does not happen all the time, as it is considered a strong augmentation.
  \item $\mathbf{N}_{basket}$: Number of images to sample into a basket. Larger number may improve variety of instances to be pasted, but increases pre-processing time. On the other hand, smaller $\mathbf{N}_{basket}$ may preserve more context information where people of a certain appearance may tend to appear together in the same image (\eg, a skiing scene).
  \item $\mathbf{R}_{paste}$: Range of number of sampled instances to be pasted --- the actual number of pasted instances each iteration is uniformly sampled within this range. An optimal number is chosen, demonstrated in section \ref{sec:numpasted} later on. We need enough occlusion to happen, yet not too many as they may over-clutter the image, which may be detrimental to the learning. $\mathbf{N}_{basket}$ will need to increase as $\mathbf{R}_{paste}$ increases such that there are enough instances in the basket to sample from.
\end{itemize}

\noindent Already, in order to address occlusions, our basic implementation differs from SCP \cite{ghiasi2021simple}: our instances are stochastically selected from a dynamic basket and pasted at random locations, while SCP rigidly takes instances from one image and directly overlays onto another at fixed original positions.

\subsection{Occlusion Copy \& Paste}

Various improvements are then made on top of our Basic Copy \& Paste pipeline to improve efficiency and introduce more variety in augmentation, further tailoring it to tackle occlusion. \emph{More details on these add-ons can be found in the supplementary material.}

\begin{wrapfigure}{r}{0.4\textwidth}
\vspace{-5mm}
\centering
\includegraphics[width=0.4\textwidth]{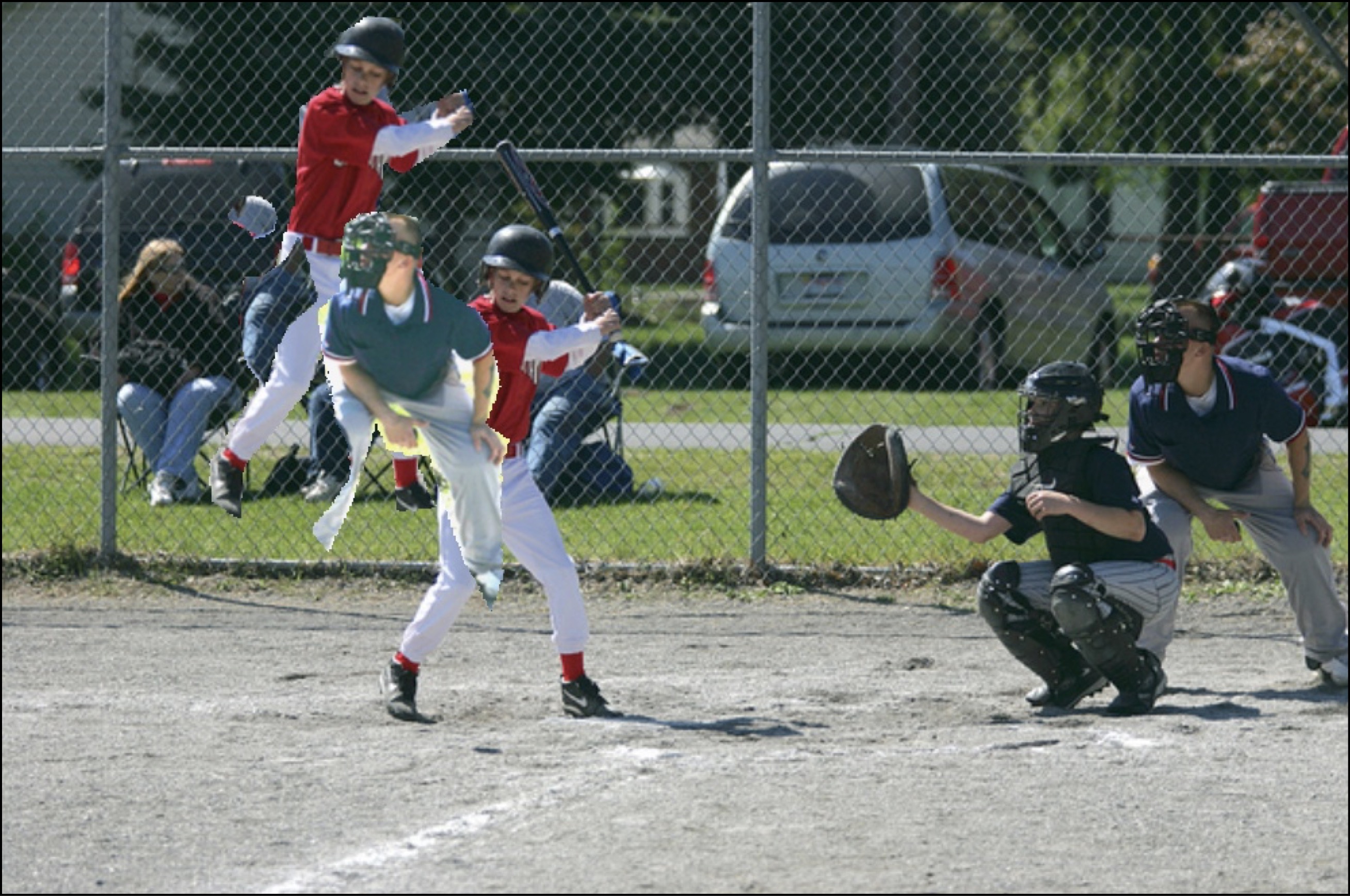}
\vspace*{-5mm}
\caption{Example of targeted pasting and augmented instance pasting. Pasted instances sampled within image to show contrast after instance augmentation}\label{fig:targeted_aug}
\vspace{-5mm}
\end{wrapfigure}
\vspace{-3mm}
\paragraph{Targeted Pasting.}\label{par:targeted} Targeted pasting works to increase efficiency of the copy \& paste augmentation to more directly induce occlusion. In target pasting, the pasting location of each copied instance is randomly chosen within the locality of a random existing instance in the image (previously, each instance is randomly pasted within the image bounds).
\vspace{-4mm}
\paragraph{Augmented Instance Pasting.}\label{par:augpaste} Augmented instance pasting works to introduce more variety in pasting instances during training. Augmentation on instances consist of a mixture of random color jittering (saturation, contrast, brightness and sharpness) and geometric jittering (scaling and rotation). 
\vspace{-2mm}

\subsection{Realism Enhancements}
\label{sec:realism}
As with all kinds of synthetic approaches, a natural question to ask is how realistic are the synthesized images for our model to learn from and generalize to real-world testing. An image’s realism depends on various factors: In synthesis via graphics rendering, photo realism is the main factor people care about. However, the very advantage of the copy \& paste approach is we get around photo realism by simply re-using parts of other real images. On the other hand, semantic realism is relatively more lacking, as pasted instances often look out of place due to the random pasting. We briefly describe some realism enhancements implemented for experimentation. It is not proven that we need \emph{perfect} realism to train a good model --- we will show in later sections that most of these realism enhancements may actually be counter-productive. Furthermore, in single class setting like this, there is an even lesser need to learn inter-class contextual dependencies (classes that naturally occur more often with each other). \emph{Please refer to supplementary for more visualised examples.}
\vspace{-4mm}
\paragraph{Minimum size filter.} We may end up with pasted instances that are too small in the augmented image. Sometimes, tiny instances are annotated and can be predicted by model because of contextual information, for \eg, if an individual person is picked out from the audience stand of a baseball game and pasted elsewhere, it may not be reasonable for any model nor humans to detect the instance if it is too small. We exclude tiny instances for pasting thresholded on the ratio of equalized side length ($\sqrt{area}$) of instance to target image.
\vspace{-8mm}
\paragraph{Scale-aware Pasting.} To improve semantic realism, in scale-aware pasting, we obtain the size distribution of the bounding boxes of original instances within the destination image, and subsequently sample a scale from this distribution for each instance to be pasted in. This way, pasted instances will look more natural and aligned with context in the scene (for \eg, wide shot or close-up shot of a scene, \emph{see supplementary}).
\vspace{-4mm}
\paragraph{Better quality masks.} \label{par:lvis} We experiment if better quality masks for copy \& paste lead to better training. We tap on the \emph{LVIS} \cite{gupta2019lvis} dataset, which uses MS COCO \cite{lin2014microsoft} images but provides better quality mask labels. However, as \emph{LVIS} is labelled in a federated style, it contains less labelled instances in less images: 15,482 person instances in 2,225 images, compared to 262,465 in 64,115 images in COCO.  
\vspace{-4mm}
\paragraph{Blending.} In \cite{dwibedi2017cut}, blending aims to remove boundary artifacts after pasting. Similarly, we implemented Gaussian blurring that operates on the pasting masks. The size and standard deviation of the Gaussian kernel controls how much ``blend'' we introduce. These hyper-parameters can be randomized at every paste during training (known as ``random blending”).

\section{Experiments}
\label{sec:experiments}

\subsection{Dataset and Evaluation Protocol}

\paragraph{Datasets.} In all experiments, we train on the \emph{person} class subset of the MS COCO dataset \cite{lin2014microsoft}, with 262,465 human instances in 64,115 images. Our copy \& paste augmentation taps on the same training set. Only for the experiment for better quality masks (as described in \cref{par:lvis}), we additionally utilize \emph{LVIS} \cite{gupta2019lvis} mask annotations. To evaluate performance of instance segmentation under highly occluded scenarios, we test on the \emph{OCHuman} \cite{zhang2019pose2seg} validation (4,291 instances in 2,500 images) \& test set (3,819 instances in 2,231 images). 
\paragraph{Fully Labelled OCHuman.} It is found that the \emph{OCHuman} benchmark is not exhaustively annotated: there are some human instances that do not have mask labels. This led to reproducibility and evaluation fairness issues which are \emph{further discussed in our supplementary material}. For fairer evaluation, we provide a subset of \emph{OCHuman} which only contains images that are fully labelled (especially important for reproduction of Pose2Seg results in \Cref{tab:sota}). This subset, \emph{OCHuman\textsuperscript{FL}} has 2,240 instances in 1,113 images (in val) and 1,923 instances in 951 images (in test). 
We report our main results on both sets: \emph{OCHuman} and \emph{OCHuman\textsuperscript{FL}}. We hope \emph{OCHuman\textsuperscript{FL}} can re-anchor a new starting point for future instance segmentation works to benchmark upon for fair evaluation and comparison.
\paragraph{Evaluation.} We report the segmentation mean average precision ($mAP$), standard evaluation protocol for instance segmentation \cite{lin2014microsoft}, equivalent to $AP$ in a one-class setting.

\subsection{Implementation details}

For all experiments (less \cref{sec:sota}), we use the same model for consistency: Mask R-CNN \cite{he2017mask} with ResNet-50 \cite{he2016deep} backbone \& Feature Pyramid Network \cite{lin2017feature}, considered a strong baseline for instance segmentation and strikes a good balance between accuracy \& training speed / GPU memory. \cite{he2019rethinking, ghiasi2021simple} have shown that when training with strong augmentations, initializing with pre-trained ImageNet \cite{deng2009imagenet} weights might actually hurt performance. We notice the same from our initial experiments, and therefore we train our whole model from scratch: convolutional weights are initialized with Kaiming initialization \cite{he2015delving} and batch normalization trainable parameters are initialized to 1. We train on single machines with 4 NVIDIA V100 GPUs each, with Synchronized Batch Normalization \cite{peng2018megdet}. We mostly follow training hyper-parameters in \cite{aggarwal_wu_lo_girshick_2021}: an initial learning rate of 0.0125 at a batch size of 8 (scaled according to the linear scaling rule \cite{goyal2017accurate}), momentum of \num{0.9} and weight decay of \num{4e-5}. We train for 75 epochs (longer due to training from scratch) with learning rate decay by a factor of \num{0.1} on the \nth{66} \& \nth{72} epochs. We use MMDetection \cite{chen2019mmdetection} to perform our experiments, and the pre-trained models which we compare against are also from MMDetection.

\vspace{-2mm}

\subsection{Baseline results}
\label{sec:baseres}

Firstly, we show on \Cref{tab:baseline} that just training with our Basic Copy \& Paste augmentation, significant improvements over the pre-trained model \& baseline vanilla trained model are evident. With the same training settings as the baseline vanilla training, the model trained additionally with Basic Copy \& Paste outperforms across the board. Additionally, $AP_{person}$ on COCO  (mini 5k) validation set is also reported here to show that in getting better at picking out occluded people, we do not sacrifice on the accuracy in normal cases.

\begin{table}[!htbp]
\vspace{-1mm}
\centering
\begin{tabular}{l c c !{\qquad} c c !{\qquad} c }
\toprule
\multirow{2}{*}{\textbf{Training Approach}} & \multicolumn{2}{c}{\textbf{OCHuman}} & \multicolumn{2}{c}{\textbf{OCHuman\textsuperscript{FL}}} & \multicolumn{1}{c}{\textbf{COCO}} \\
 & \multicolumn{1}{c}{$\bm{AP^{val}}$} & \multicolumn{1}{c}{$\bm{AP^{test}}$} & \multicolumn{1}{c}{$\bm{AP^{val}}$} & \multicolumn{1}{c}{$\bm{AP^{test}}$} & $\bm{AP^{val}_{person}}$  \\
\cmidrule(lr){1-1}
\cmidrule(lr){2-3}
\cmidrule(lr){4-5}
\cmidrule(lr){6-6}
Pre-trained from \cite{chen2019mmdetection} & 14.9 & \multicolumn{1}{c}{14.9} & 24.5 & \multicolumn{1}{c}{24.9} & 47.5 \\
Baseline vanilla training & 16.5 & \multicolumn{1}{c}{16.6} & 27.0 & \multicolumn{1}{c}{27.4} & 48.7 \\ 
\hdashline\noalign{\vskip 0.5ex}
\rowcolor{LightCyan}
\textbf{+ Basic Copy \& Paste (ours)} & \textbf{18.6} & \multicolumn{1}{c}{\textbf{17.8}} & \textbf{29.3} & \multicolumn{1}{c}{\textbf{28.5}} & \textbf{49.2} \\
\bottomrule
\end{tabular}
\caption{Baseline comparisons with Basic Copy \& Paste, tested on \emph{OCHuman}}
\label{tab:baseline}
\end{table}

\vspace{-3mm}
\noindent Further improvements are made after the ablation study in \cref{sec:ablation} and the \emph{final results} are reported in \cref{sec:sota}. 
\vspace{-3mm}

\subsection{Ablation study}
\label{sec:ablation}

\begin{wrapfigure}{r}{0.5\textwidth}
\vspace{-13mm}
\centering
\includegraphics[width=0.5\textwidth]{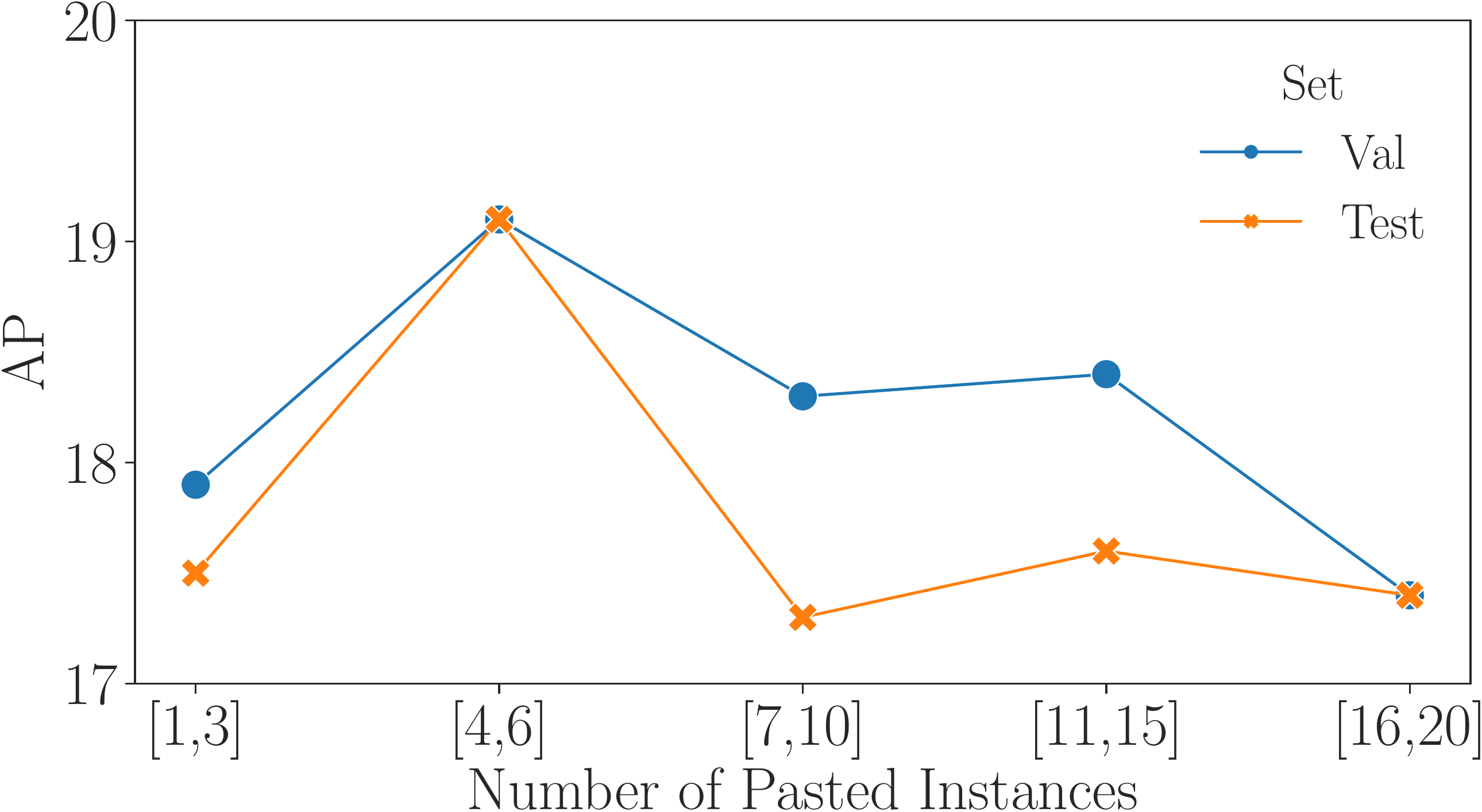}
\vspace*{-8mm}
\caption{Effect of number of pasted instances on $AP$, tested on \emph{OCHuman}}\label{fig:numpaste}
\vspace{-12mm}
\end{wrapfigure}

In this section, we step through experiments in our ablation study to further tailor our eventual Occlusion Copy \& Paste for the best performance.

\subsubsection{Number of pasted instances}
\label{sec:numpasted}
\vspace{-1mm}
We study how number of pasted instances affect the performance of the trained model. The results are plotted out on \Cref{fig:numpaste}. When there are too little pasted instances, there is little chance of occlusion occurring which contributes to poorer performance. On the other hand, when there are too many pasted instances, performance drops possibly due to over-cluttering affecting learning. Optimal $R_{paste}$ is $\interval{4}{6}$ for Basic Copy \& Paste.

\vspace{1mm}

\begin{table}[!htbp]
  \centering
  \small
  \begin{tabular}{l c l c l}
    \textbf{copy \& paste add-ons}  & \bm{$AP^{val}$} & & \bm{$AP^{test}$} & \\
    \Xhline{3\arrayrulewidth}
    Basic Copy \& Paste             & 18.6 & & 17.8 & \\
    Minimum Pasting Size                        & \textbf{18.8} & & \textbf{18.5} & \\ \hdashline
    Minimum Pasting Size + Scale Aware          & 18.2 & \textcolor{red}{-0.6} & 18.2 & \textcolor{red}{-0.3} \\
    Minimum Pasting Size + Better Quality Mask  & 18.4 & \textcolor{red}{-0.4} & 18.0 & \textcolor{red}{-0.5} \\
    Minimum Pasting Size + Blend (Fixed)        & 18.6 & \textcolor{red}{-0.2} & 17.9 & \textcolor{red}{-0.6} \\
    Minimum Pasting Size + Blend (Random)       & \textbf{19.0} & \textcolor{green}{+0.2} & 18.4 & \textcolor{red}{-0.1} \\
  \end{tabular}
  \caption{Ablation on realism enhancers, tested on \emph{OCHuman}}
  \label{tab:ablation_realism}
  \vspace{-8mm}
\end{table}

\subsubsection{Realism enhancement experiments}
Next, we experimented on the effects of the various realism enhancements described in section \ref{sec:realism}. As seen on \Cref{tab:ablation_realism}, imposing minimum pasting size led to a slight improvement --- not pasting tiny instances which are not meaningful for the model to predict and learn from, does help the model to learn slightly better. However, beyond that, we see that in general, the rest of the realism enhancements actually caused performance to drop. The argument against realism enhancements is the trade-off with the scope of augmentation variability. Implementing realism means imposing a more restrictive scope on how varied pasting can be: Scale-aware pasting restricts variability on sizes of pasted instances. Using better quality masks may actually help, but in this case, using \emph{LVIS} for better quality masks means fewer human instances to copy from. Variability of pasted instances is more important than quality of instances. Random blending does perform slightly better on the validation set, but the slight improvement does not justify the increase of about 20\% in training time due to significant computation in blending operations. Our findings that realism enhancements are generally counter-productive corroborates with \cite{ghiasi2021simple}. Moving on, we only preserve the minimum pasting size control for Occlusion Copy \& Paste.

\subsubsection{Efficacy of targeted pasting \& instance augmentation}
One potential downside of Basic Copy \& Paste was that training time increases with a greater number of pasted instance due to the sequential pasting of instances. This downside is minimized with targeted pasting (described in \cref{par:targeted}). As seen on \Cref{tab:ablation_targeted_aug}, targeted pasting improves the efficiency of pasting such that we achieve the very good performance at low $R_{paste}$ of $\interval{1}{3}$. In fact, at higher pasting number with targeted pasting, performance improvements are not as obvious and even starts to drop, possibly due to over-cluttering around a local region. Targeted pasting allows us to achieve high performance with less pasted instances and consequentially, shorter training duration. We also observe that augmented pasting (described in \cref{par:augpaste}) further pushes the performance on \emph{OCHuman}. Once again, this demonstrates that more variability in augmentation helps with performance. 

\begin{table}[!htbp]
  \centering
  \small
  \begin{tabular}{l c l c l}
    \textbf{copy \& paste add-ons} & \bm{$AP^{val}$} & & \bm{$AP^{test}$} & \\
    \Xhline{3\arrayrulewidth}
    Basic C\&P, $R_{paste}=\interval{1}{10}$     & 18.6 & & 17.8 & \\
    + Targeted                       & 18.6 & \textcolor{green}{+0.0} & 18.2 & \textcolor{green}{+0.4} \\ 
    
    \Xhline{2\arrayrulewidth}

    Basic C\&P, $R_{paste}=\interval{1}{3}$      & 17.9 & & 17.5 & \\
    + Targeted                       & 19.1 & \textcolor{green}{+1.2}  & 18.0 & \textcolor{green}{+0.5} \\ 
    + Targeted \& Augm. Paste     & 19.2 & \textcolor{green}{+1.3} & 18.4 & \textcolor{green}{+0.9} \\
    \textbf{+ Targeted, Augm. Paste \& Min. Size}     & \textbf{19.5} & \textbf{\textcolor{green}{+1.6}} & \textbf{18.6} & \textbf{\textcolor{green}{+1.1}} \\
  \end{tabular}
  \caption{Ablation on Targeted \& Augmented Paste, tested on \emph{OCHuman}}
  \label{tab:ablation_targeted_aug}
\end{table}

\noindent Our eventual Occlusion Copy \& Paste (OC\&P) is made out of the Basic Copy \& Paste with minimum pasting size control imposed, targeted pasting at a $R_{paste}$ of $\interval{1}{3}$ and augmented instance pasting. 

\vspace{-2mm}
\subsection{Pushing the performance \& SOTA}
\label{sec:sota}
On \Cref{tab:sota}, we show that OC\&P is easily interoperable with any model: besides Mask R-CNN (previously in \cref{sec:baseres,sec:ablation}), we trained OC\&P with Pose2Seg \cite{zhang2019pose2seg} and showed a significant $\sim$10\% improvement on \emph{OCHuman\textsuperscript{FL}}. Finally, we applied OC\&P on Mask2Former (M2F) \cite{cheng2022masked}, one of vastly different model architecture (see \cref{par:related_od}), and push the SOTA performance on instance segmentation task on the \emph{OCHuman} benchmark. We use M2F with Swin-S \cite{liu2022swin} backbone, follow their original training schemes and fine-tune on our COCO training set --- this includes Large Scale Jittering (LSJ) augmentation, which is a strong augmentation first introduced in \cite{ghiasi2021simple}, which does random image resizing at a larger range of $\interval{0.1}{2.0}$, followed by fixed size cropping. This demonstrates that OC\&P is also additive to other strong augmentation methods. \emph{More details on our M2F experiments can be found in supplementary}. As seen on the last row of \Cref{tab:sota}, our eventual model trained with OC\&P outperforms PoSeg \cite{zhou2020poseg}, the current SOTA on \emph{OCHuman}. The performance on \emph{OCHuman\textsuperscript{FL}} doubles that of Pose2Seg \cite{zhang2019pose2seg} as well. The model architectures of Pose2Seg \& PoSeg were specially designed to tackle occlusions, and ExPoSeg even utilises a high-performance external pose estimation model. M2F is a generic instance segmentation model, but is able to outperform both models just by training with OC\&P --- this shows the potential of such data-centric approaches. For completeness of comparison, we also include results from our training runs with Simple Copy-Paste \cite{ghiasi2021simple} (details in \cref{par:related_cp}) instead of OC\&P and show that OC\&P still outperforms the simpler augmentation method. 

\begin{table}[!htbp]
\vspace{-1mm}
\centering
\begin{threeparttable}
\begin{tabular}{l c@{\hskip 0mm} c c c c c}
\toprule
\multirow{2}{*}{\textbf{Model}} & \multirow{2}{15mm}{\centering \scriptsize \textbf{External\\Pose Model}} & \multirow{2}{15mm}{\centering \scriptsize \textbf{Modelled for\\Occlusion}} & \multicolumn{2}{c}{\textbf{OCHuman}} & \multicolumn{2}{c}{\textbf{OCHuman\textsuperscript{FL}}} \\
  &  &   & \multicolumn{1}{c}{$\bm{AP^{val}}$} & \multicolumn{1}{c}{$\bm{AP^{test}}$} & \multicolumn{1}{c}{$\bm{AP^{val}}$} & \multicolumn{1}{c}{$\bm{AP^{test}}$} \\
\cmidrule(lr){1-1} \cmidrule(lr){2-3} \cmidrule(lr){4-5} \cmidrule(lr){6-7}
Pose2Seg$^\mathsection$ \cite{zhang2019pose2seg} & &  & - & \multicolumn{1}{c}{-} & 22.8\tnote{+} & 22.9\tnote{+}\\
\rowcolor{LightCyan}
\emph{+ Occlusion C\&P (ours)} & \multirow{-2}{*}{\LARGE \cmark} & \multirow{-2}{*}{\LARGE \cmark} & - & \multicolumn{1}{c}{-} & \underline{25.3}\tnote{+} & \underline{25.1}\tnote{+} \\
\hdashline\noalign{\vskip 0.5ex}
Mask R-CNN$^\mathsection$ \cite{he2017mask} & \multirow{3}{*}{\huge \xmark} & \multirow{3}{*}{\huge \xmark} & 14.9 & \multicolumn{1}{c}{14.9} & 24.5 & 24.9 \\
Mask R-CNN$^\dagger$ & & & 16.5 & \multicolumn{1}{c}{16.6} & 27.0 & 27.4 \\
\rowcolor{LightCyan}
\emph{+ Occlusion C\&P (ours)} & & & \underline{19.5} & \multicolumn{1}{c}{\underline{18.6}} & \underline{30.6} & \underline{29.9} \\
\hdashline\noalign{\vskip 0.5ex}
PoSeg (JoPoSeg) \cite{zhou2020poseg} & \small \xmark & \multirow{2}{*}{\LARGE \cmark} & 25.8\tnote{$\star$} & \multicolumn{1}{c}{26.4\tnote{$\star$}} & - & - \\
PoSeg (ExPoSeg) & {\small \cmark} & & 26.4\tnote{$\star$} & \multicolumn{1}{c}{26.8\tnote{$\star$}} & - & - \\
\hdashline\noalign{\vskip 0.5ex}
Mask2Former$^\mathsection$ \cite{cheng2022masked} & & & 25.9 & \multicolumn{1}{c}{25.4} & 43.2 & 44.7 \\
Mask2Former$^\dagger$ & & & 26.7 & \multicolumn{1}{c}{26.3} & 45.2 & 46.4 \\
+ Simple Copy-Paste \cite{ghiasi2021simple} & & & 28.0 &  \multicolumn{1}{c}{27.7} & 48.9 & 50.2 \\
\rowcolor{LightCyan}
\textbf{+ Occlusion C\&P (ours)} & \multirow{-4}{*}{\Huge \xmark} & \multirow{-4}{*}{\Huge \xmark} & \textbf{28.9} & \multicolumn{1}{c}{\textbf{28.3}} & \textbf{49.3} & \textbf{50.6} \\
\bottomrule
\end{tabular}
\vspace{-4mm}
\caption{OC\&P improves $AP$ across the board and achieves SOTA performance on \emph{OCHuman}. $\bm{|}$ $\mathsection$: pre-trained models $\bm{|}$ $\dagger$: models from our baseline vanilla training}
\begin{tablenotes}
\small
\item[$\star$] Directly referenced from paper as no code is published
\item[+] An exhaustively-labelled \emph{OCHuman\textsuperscript{FL}} is important for fair evaluation in Pose2Seg. Evaluating on \emph{OCHuman\textsuperscript{FL}} allow us to closely reproduce results reported in \cite{zhang2019pose2seg}. 
\end{tablenotes}
\label{tab:sota}
\vspace{-5mm}
\end{threeparttable}
\end{table}

\vspace{-1mm}
\section{Conclusion}
\label{sec:conclusion}
\vspace{-2mm}
All in all, we propose a novel use of copy \& paste augmentation to tackle the difficult problem of same-class occlusion in instance segmentation. From Basic Copy \& Paste, we experimented with various add-ons: we found realism enhancements are mostly counter-productive, but targeted pasting \& augmented pasting improves performance through increased efficiency and variability in augmentation. The eventual Occlusion Copy \& Paste augmentation takes these elements and we show that it is interoperable with SOTA instance segmentation models, significantly improving performance on occluded scenarios for “free”, without any additional data or labels. Even without explicit architectural design to tackle occlusions, we outperform the SOTA on \emph{OCHuman} by simply applying our Occlusion Copy \& Paste on a generic SOTA instance segmentation model. This demonstrates the potential of data-centric approaches. A key benefit of our approach is that it is easily applied with any models or other model-centric improvements. Given the speed at which the deep learning field moves, it is to everyone’s advantage to have approaches that are highly interoperable with every other aspect of training. We leave as future work to integrate this with model-centric improvements to effectively solve occluded person instance segmentation.

\newpage
\bibliography{egbib}
\end{document}